\documentclass[runningheads]{llncs}
\usepackage{graphicx}
\usepackage{comment}
\usepackage{amsmath,amssymb}
\usepackage{color}
\usepackage{wrapfig}
\usepackage{marvosym}

\begin{document}
\pagestyle{headings}
\mainmatter
\def\ECCVSubNumber{3124}  

\title{Deep Complementary Joint Model for Complex Scene Registration and Few-shot Segmentation on Medical Images} 

\titlerunning{Deep Complementary Joint Model}
%
\author{Yuting He\inst{1}\orcidID{0000-0003-0878-8915} \and
Tiantian Li\inst{1}\and
Guanyu Yang\inst{1,2}\textsuperscript{(\Letter)}\orcidID{0000-0003-3704-1722}\and
Youyong Kong\inst{1,2}\and
Yang Chen\inst{1,2}\and
Huazhong Shu\inst{1,2}\and
Jean-Louis Coatrieux\inst{2,3}\and
Jean-Louis Dillenseger\inst{2,3}\and
Shuo Li\inst{4}\orcidID{0000-0002-5184-3230}}
\authorrunning{Y. He et al.}
%
\institute{Laboratory of Image Science and Technology, Southeast University, Nanjing 210096, China\\
\email{yang.list@seu.edu.cn} (G. Yang\textsuperscript{(\Letter)}) \and
Centre de Recherche en Information Biomédicale Sino-Français (CRIBs)\and
Univ Rennes, Inserm, LTSI - UMR1099, Rennes, F-35000, France \and
Dept. of Medical Biophysics, University of Western Ontario, London, ON, Canada \\
\email{slishuo@gmail.com} (S. Li)}
\maketitle

\begin{abstract}
Deep learning-based medical image registration and segmentation joint models utilize the complementarity (augmentation data or weakly supervised data from registration, region constraints from segmentation) to bring mutual improvement in complex scene and few-shot situation. However, further adoption of the joint models are hindered: 1) the diversity of augmentation data is reduced limiting the further enhancement of segmentation, 2) misaligned regions in weakly supervised data disturb the training process, 3) lack of label-based region constraints in few-shot situation limits the registration performance. We propose a novel Deep Complementary Joint Model (DeepRS) for complex scene registration and few-shot segmentation. We embed a perturbation factor in the registration to increase the activity of deformation thus maintaining the augmentation data diversity. We take a pixel-wise discriminator to extract alignment confidence maps which highlight aligned regions in weakly supervised data so the misaligned regions' disturbance will be suppressed via weighting. The outputs from segmentation model are utilized to implement deep-based region constraints thus relieving the label requirements and bringing fine registration. Extensive experiments on the CT dataset of MM-WHS 2017 Challenge\cite{zhuang2016multi} show great advantages of our DeepRS that outperforms the existing state-of-the-art models.
\end{abstract} 

\section{Introduction}
Deep learning-based medical image segmentation models and registration models \cite{litjens2017survey,haskins2019deep,lateef2019survey} are limited in complex scene and few-shot situation. In complex scene which has complex but task-unconcerned backgrounds, the unsupervised registration models \cite{balakrishnan2018unsupervised,fan2019adversarial} pay equal attention to all regions for overall alignment so that the performance on regions of interest (ROIs) will be limited by background. In few-shot situation which lacks labels, the segmentation models \cite{ronneberger2015u,long2015fully} will over-fit \cite{zhao2019data,shorten2019survey} due to the lack of supervision information.

\begin{wrapfigure}{r}{0.6\linewidth}
\centering
\includegraphics[width=\linewidth]{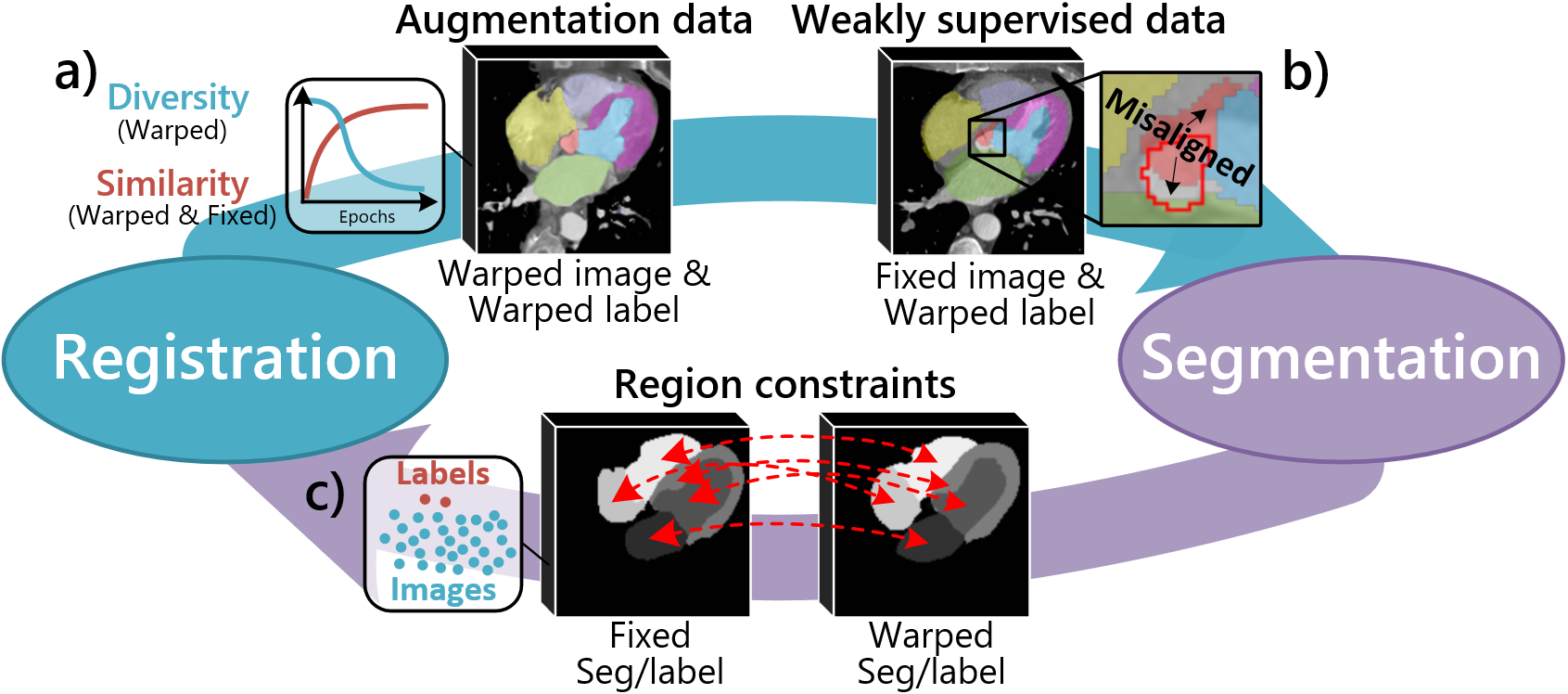}
\caption{The complementary topology and limitations of registration and segmentation tasks. Registration provides augmentation data and weakly supervised data for segmentation for higher generalization in few-shot situation, the segmentation feeds back region constraints for finer registration on ROIs in complex scene. a), b), c) illustrate the limitations in the utilization of this complementary topology.}
\label{fig:RS}
\end{wrapfigure}

The registration and segmentation tasks has great complementarity which will bring mutual improvement in complex scene and few-shot situation. As shown in Fig.~\ref{fig:RS}, the registration model provides diverse augmentation data (warped images and labels) or weakly supervised data (fixed images and warped labels) for segmentation model \cite{zhao2019data,xu2019deepatlas} during the training process, thus reducing the requirement of labels and enhancing the segmentation generalization in few-shot situation. The segmentation model feeds back region constraints \cite{li2019hybrid,xu2019deepatlas,estienne2019u} so that additional attention on ROIs is paid for finer registration in complex scene.

Unfortunately, further exploiting of this complementary topology are hindered \cite{zhao2019data,xu2019deepatlas,estienne2019u,li2019hybrid} due to: \textbf{Limitation 1:} \emph{Degradation of data augmentation capability} (Fig.~\ref{fig:RS}(a)). During the training of registration model, it learns the deformation rule that matches real situation and generates diverse warped images as augmentation data to improve the segmentation generalization ability \cite{xu2019deepatlas,de2019deep}. However, the similarity between warped and fixed images increases and tends to become stable, and the diversity of warped images is gradually reduced as the similarity stabilizes. Therefore, in the later training stage of registration network, the identical warped images are generated in different epochs, resulting in the reduction of augmentation data diversity. Thus, the data augmentation ability of registration model is degraded and the further enhancement of segmentation will be limited. \textbf{Limitation 2:} \emph{Misaligned regions in weakly supervised data} (Fig.~\ref{fig:RS}(b)). The weakly supervised data enlarges the labeled dataset and provide additional supervision information for the segmentation model. However, large misaligned regions in these data will produce incorrect optimization targets and it will disturb the training process leading to serious mis-segmentation if used directly \cite{xu2019deepatlas}. \textbf{Limitation 3:} \emph{Lack of label-based region constraints} (Fig.~\ref{fig:RS}(c)). Region constraints provide specific alignment information for regions bringing finer registration optimization. However, in few-shot situation, the label-based region constraints \cite{xu2019deepatlas,estienne2019u,li2019hybrid,hu2018weakly} are lacked with few labels. Thus if in complex scene, the registration model \cite{balakrishnan2018unsupervised,de2019deep,fan2019adversarial} will take rough optimization and the complex backgrounds will limit the registration performance on ROIs.

\textbf{Solution 1} for the degradation of data augmentation capability: we embed a random perturbation factor in the registration to increase the activity of deformation for sustainable data augmentation capability. The registration process is a displacement of structure information, and the adjustment of deformation degree is the sampling of the structure information on this displacement path \cite{learned2005data,hauberg2016dreaming}. Therefore, our perturbation factor adjusts the deformation degree randomly to sample the structure information which is consistent with the real distribution to produce diverse and real augmentation data for the segmentation model.

\textbf{Solution 2} to suppress the misaligned regions' disturbance: we extract alignment confidence maps from a pixel-wise discriminator to suppress the misaligned regions in weakly supervised data and utilize the supervision information in aligned regions. The pixel-wise discriminator, resulting in a generative adversarial network (GAN) \cite{goodfellow2014generative} based registration model \cite{fan2019adversarial,yan2018adversarial,fan2018adversarial,hu2018adversarial}, learns the similarity between warped and fixed images and outputs the alignment confidence maps that highlight the aligned regions \cite{hung2018adversarial,nie2018asdnet}. Thus, via these maps ,the misaligned regions will be suppressed and the supervision information in aligned regions will be utilized for higher segmentation generalization when calculating the weakly supervised loss function.

\textbf{Solution 3} to cope with the lack of label-based region constraints: we build deep-based region constraints that calculate the loss value via the warped and fixed segmentations from the segmentation model so that fine registration optimization targets are available. Therefore, 1) label requirements of label-based region constraints are freed in few-shot situation, 2) different regions are independently optimized to avoid the misalignment of each region and 3) region attention on the ROIs is paid for finer registration.

In this paper, we propose a \emph{Deep Complementary Joint Model (DeepRS)} that minimizes background interference in complex scene for finer registration on ROIs, and greatly reduces the label requirements of segmentation in few-shot situation for higher generalization ability. In short, the contributions of our work are summarized as follows:
\begin{itemize}
\item To the best of our knowledge, we build a novel complementary topology of registration and segmentation for the first time, and propose the DeepRS model utilizing the data generation ability of registration for few-shot segmentation, and the label-free region constraint ability of segmentation for complex scene registration.

\item We propose a deep structure sampling (DSS) block adding a random perturbation factor to the registration for sustainable data augmentation ability.

\item We propose an alignment confidence map (ACM) method which efficiently utilizes the supervision information in weakly supervised data thus bringing powerful segmentation generalization.

\item We propose a deep-based region constraint (DRC) strategy which frees up the label requirements of label-based methods achieving finer registration on ROIs.
\end{itemize}

\section{Related Works}
\subsection{Registration and segmentation joint models}
Registration and segmentation tasks have great complementarity, thus building a registration and segmentation joint model has the potential of mutual improvement. The registration provides augmentation data and weakly supervised data for the segmentation \cite{zhao2019data,xu2019deepatlas,vakalopoulou2018atlasnet}, and the segmentation feeds back additional region constraints \cite{hu2018weakly,li2019hybrid,estienne2019u}. Zhao \emph{et al.} \cite{zhao2019data} took a pre-trained registration model to generate augmentation data for more powerful segmentation ability. Li \emph{et al.} \cite{li2019hybrid} made a hybrid framework that took the label-based region constraints from labels and segmentations for finer registration. Similarity, Xu \emph{et al.} \cite{xu2019deepatlas} designed a semi-supervised method that combined registration and segmentation models bringing the mutual improvement in knee and brain images.

However, these existing methods only took the advantage of partial complementarity which hardly gives full play to their potential. The convergence of the registration model limits the diversity of the augmentation data and prevents further enhancement of the segmentation model \cite{zhao2019data}. Misaligned regions in weakly supervised data disturb the training of segmentation models, and if used directly, it will lead to serious mis-segmentation \cite{xu2019deepatlas}. In few-shot situation, label-based region constraints are lacked due to the small labeled dataset \cite{li2019hybrid}, thus with inaccurate optimization targets, complex backgrounds will limit registration performance on ROIs in complex scene.
\subsection{Data augmentation}
Data augmentation \cite{shorten2019survey}, generating bigger dataset, has the ability to improve learning models\cite{Sun_2017_ICCV}, especially in few-shot situation. Some data augmentation strategies (random cropping, mirroring, rotation, flipping, etc.) are often used for higher generation ability, while inappropriate strategy combinations will generate unreasonable data which will weaken the model performance \cite{shorten2019survey}. Learning-based data augmentation strategies \cite{cubuk2019autoaugment,nielsen2019gan,jackson2018style,lemley2017smart} learn the augmentation methods from dataset for real augmentation data. Registration learns transformation rules of structure information from the images \cite{zhao2019data,learned2005data,hauberg2016dreaming,xu2019deepatlas} so that the augmentation images with real structure information are obtained.

Disappointingly, the registration-based augmentation ability will degrade due to the reduction of deformation diversity. As the registration model converges, the moving image is stably aligned onto the fixed image and the identical warped images in different epochs are generated, resulting in the reduction of augmentation data diversity and limiting the further improvement of segmentation.
\subsection{Weakly-supervised learning}
Weakly-supervised learning \cite{hu2018weakly,tang2018regularized,kolesnikov2016seed,papandreou2015weakly,hung2018adversarial,nie2018asdnet} utilizing non-precisely labeled data is a strategy for labeled data limitation. It has three typical types according to the weakly supervised data types \cite{zhou2017brief}: 1) incomplete supervision where part of the dataset without labels\cite{nie2018asdnet,hung2018adversarial}, 2) inexact supervision where data with coarse-grained labels \cite{hu2018weakly,kolesnikov2016seed} and 3) inaccurate supervision where data with inaccurate labels \cite{tang2018regularized,tang2018normalized}. In registration and segmentation tasks, the warped labels and fixed images from registration model make up weakly supervised data leading to inaccurate supervision which will improve the segmentation performance with appropriate strategy. Unfortunately, if the weakly supervised data is used directly, the misaligned regions will brings inaccurate optimization target, thus disturbing the training process and lead to mis-segmentation.
\begin{wrapfigure}{r}{0.6\linewidth}
\begin{center}
\includegraphics[width=\linewidth]{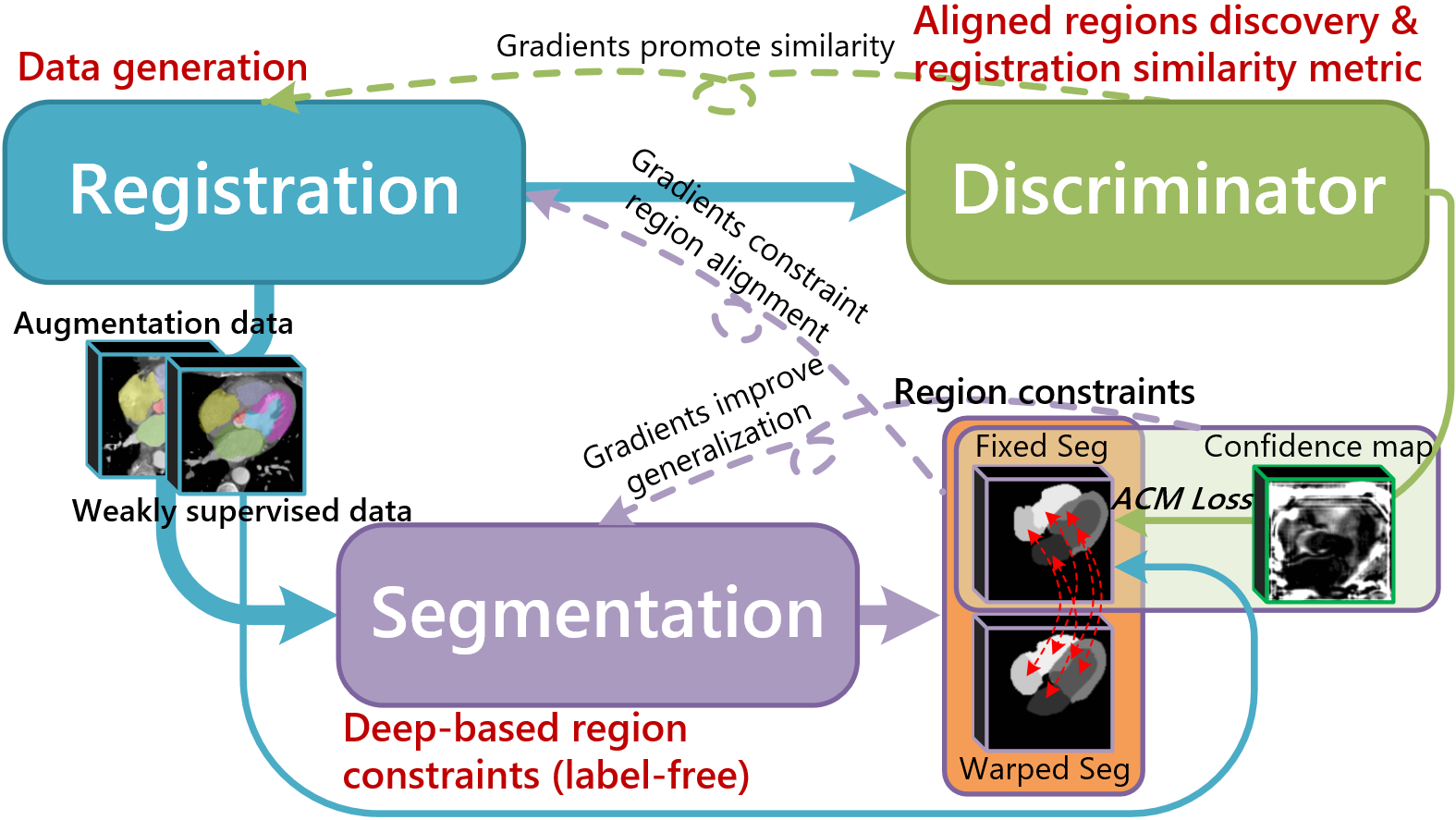}
\end{center}
\caption{The overview of our DeepRS. The data generation ability of the registration, the deep-based region constraint of the segmentation, the aligned regions discovery ability and the learned similarity metric of discriminator interact in the alternating training process.}
\label{fig:overview}
\end{wrapfigure}
\subsection{Generative Adversarial Networks}
Generative adversarial networks (GANs) \cite{goodfellow2014generative,yi2019generative,ge2019stereo}, consisting of a generator $G$ and a discriminator $D$, learns a similarity metric of the generated and real images. The discriminator learns to distinguish the real or generated images and the generator takes the adversarial loss from the discriminator to improve the authenticity of the generated image to deceive the discriminator. GAN-based registration models \cite{fan2019adversarial,yan2018adversarial,fan2018adversarial,hu2018adversarial} take global discriminator to learn image-wise similarity metric of warped and fixed images which can be used to evaluate the weakly supervised data in our task.

However, the image-wise similarity has no ability to evaluate the regional similarity and in our segmentation task (pixel-wise), it will still introduce the error information in the weakly supervised data. Patch-GANs utilize pixel-wise discriminator \cite{nie2018asdnet,hung2018adversarial} consisting of a full convolution network to learn the pixel-wise similarity and output confidence maps which highlight task-beneficial regions. Thus, a patch-GAN is used in our model for alignment confidence maps to suppress the misaligned regions and utilize the supervision information in weakly supervised data.

\section{Methodology}
\begin{figure*}
\begin{center}
\includegraphics[width=11cm]{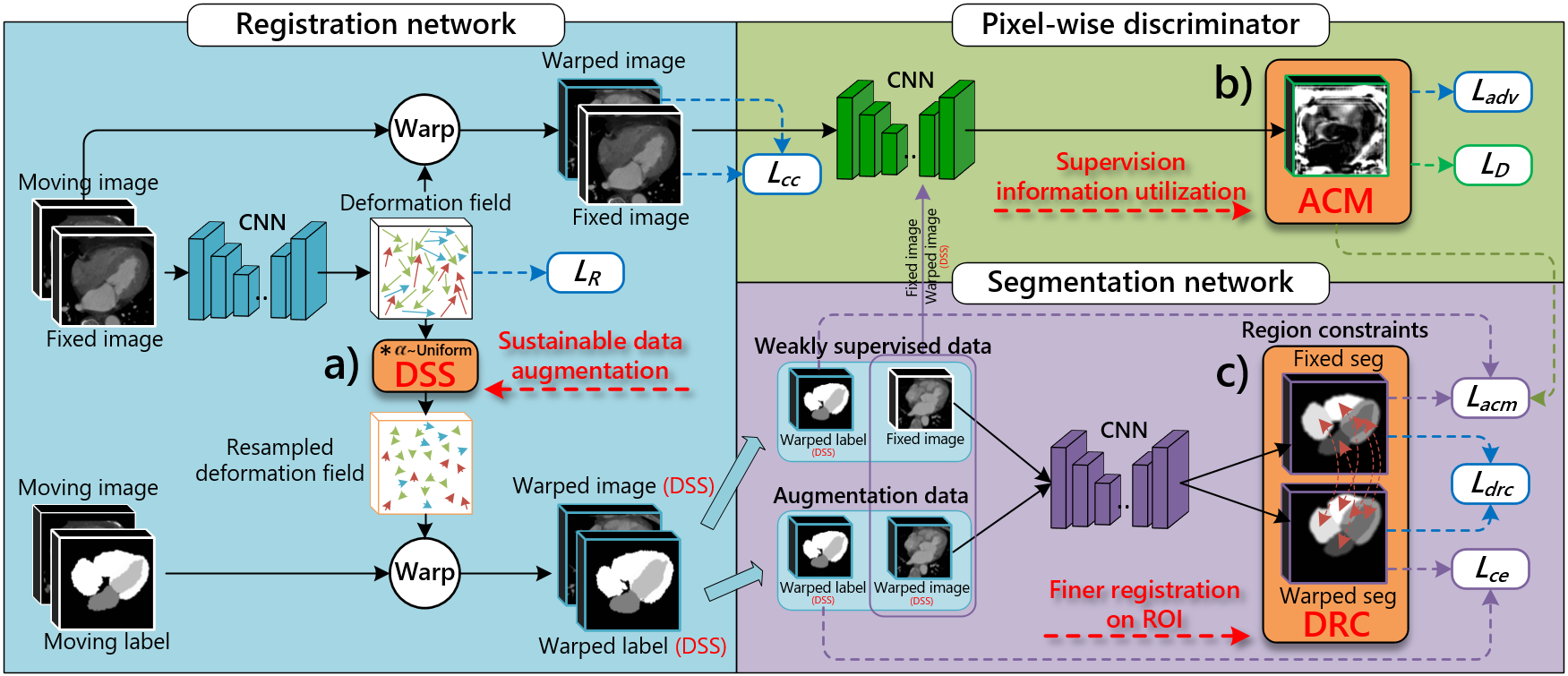}
\end{center}
\caption{In detail of our DeepRS model, we design a DSS block, a ACM method and a DRC strategy cleverly dealing with the limitations. a) The DSS block maintains the diversity of warped images bringing sustainable data augmentation ability. b) The ACM method utilizes the supervision information in weakly supervised data. c) The DRC strategy provides region attention on ROIs for finer registration.}
\label{fig:deeprs}
\end{figure*}

Our DeepRS model (Fig.~\ref{fig:deeprs}, Fig.~\ref{fig:overview}), which consists of registration, pixel-wise discriminator and segmentation models, leverages their complementarity for complex scene registration and few-shot segmentation (Sec.~\ref{sec1}) bringing mutual improvement. The registration generates diverse augmentation data via randomly adjusting the deformation field in a DSS block (Sec.~\ref{subsec1}) and provides weakly supervised data for the segmentation network to reduce the labeled data requirements in few-shot situation. The pixel-wise discriminator provides ACMs (Sec.~\ref{subsec2}) for the segmentation network for supervision information utilization in weakly supervised data. The segmentation network provides DRC (Sec.~\ref{subsec3}) for the registration network for finer registration on ROIs in complex scene. The joint strategy (Sec.~\ref{sec2}) maximizes the complementarity via alternating training.
\subsection{DeepRS for stronger registration and segmentation}
\label{sec1}
The proposed DeepRS model leverages the complementarity of registration and segmentation tasks via the DSS block, ACM method and DRC strategy.
\subsubsection{Deep structure sampling (DSS) for sustainable data augmentation}
\label{subsec1}
\begin{figure*}
\begin{center}
\includegraphics[width=11cm]{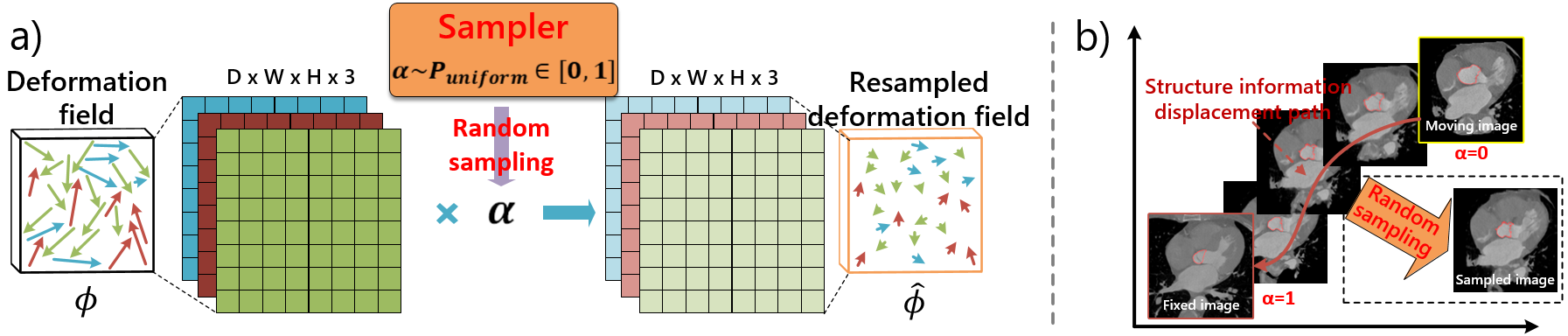}
\end{center}
\caption{The DSS block for sustainable data augmentation. a) A perturbation factor $\alpha\in[0,1]$ from uniform distribution adjusts the deformation field making the sampling process. b) Illustration of the sampling that registration makes the structure information displacement and our DSS samples the information on its displacement path.}
\label{fig:DSS}
\end{figure*}
DSS block generates diverse augmentation data sustainedly via embedding a random perturbation factor in the deformation field to increase the uncertainty of the warped images and labels. The registration process is the displacement of image structure information, and the perturbation of deformation degree realizes the sampling of information on this displacement path \cite{learned2005data,hauberg2016dreaming}. Therefore, the DSS block brings two advantages: 1) Sustainable data augmentation. The perturbation factor controls the deformation degree so that the registration network is guaranteed to generate diverse augmentation data sustainedly. 2) Real distribution. Sampling structure information from its displacement path generates the augmentation data more matching real distribution than other manual augmentation methods.

As shown in Fig.~\ref{fig:DSS}(a), a deformation field $\varnothing$ from registration network is multiplied by a random perturbation factor $\alpha$ from uniform distribution $p_{uniform}\in[0,1]$ to obtain an resampled deformation field $\hat{\varnothing}=\varnothing\times\alpha\backsim p_{uniform}\in[0,1]$. Therefore, the warped images and labels deformed by it will still have great diversity, even if the registration network has converged. Fig.~\ref{fig:DSS}(b) illustrates that as $\alpha$ increases, the warped images gradually approximate the fixed images since its structure information approaches the fixed image. It is evident that the randomly sampled deformations are non-rigid, yet produce realistically-looking images.
\subsubsection{Alignment confidence map (ACM) for supervision information utilization}
\label{subsec2}
ACM method utilizes the supervision information of aligned regions and suppresses the misaligned regions in weakly supervised data to improve the segmentation generalization ability. The ACM maps from the pixel-wise discriminator evaluate the pixel-wise similarity between warped and fixed images and will highlight the aligned regions. Thus, these maps will be taken to weight the loss of weakly supervised data to utilize its supervision information in aligned regions, as illustrated in Equ.~\ref{equ:ACM}:
\begin{equation}
   \mathcal{L}_{acm}=-D(W(x_{m},\hat{\varnothing}),x_{f})W(y_{m},\hat{\varnothing})\log{S(x_{f})}
   \label{equ:ACM}
\end{equation}
where $x_{m}$, $y_{m}$, $x_{f}$ and $\hat{\varnothing}$ are the moving image, moving label, fixed image and resampled deformation field from DSS block. As shown in Fig.~\ref{fig:deeprs}, $W(\cdot,\cdot)$ is the 'warp' block which deforms the moving images and labels to the fixed images for warped images and labels following the spatial transformation layer in \cite{balakrishnan2018unsupervised}. The pixel-wise discriminator $D(\cdot, \cdot)$ measures the similarity between warped and fixed images for the ACMs to weight the cross-entropy loss between warped labels and fixed segmentation (seg-) masks $S(x_{f})$. Therefore, the loss value in misaligned region will get low weight and the disturbance will be suppressed.

The contribution of the weakly supervised data is increasing during the training. In early training stage, the powerful discriminator outputs weak maps, so that the loss from weakly supervised data is suppressed greatly and the optimization target of the segmentation network is dominated by the loss $\mathcal{L}_{ce}$ from augmentation data. As the training progresses, the registration network defeats the discriminator and obtains high responsive maps, thus increasing the contribution of ACM loss $\mathcal{L}_{acm}$, so that the segmentation generalization ability will be further enhanced.
\subsubsection{Deep-based region constraint (DRC) for finer registration on ROIs}
\label{subsec3}
DRC strategy guides the attention on the ROIs for finer registration via constraints between the fixed and warped seg-masks from the segmentation network. This deep-based region constraint takes the alignment of the corresponding regions in warped and fixed images as the optimization target, so that 1) label requirements of label-based region constraints is freed in few-shot situation, 2) different regions are independently optimized to avoid the misalignment between each other and 3) additional region attention on the ROIs is paid for finer registration.

As shown in Fig.~\ref{fig:deeprs}(c), the warped image and the fixed image are input into the segmentation network respectively for the warped and the fixed seg-masks firstly. Then a mean square error loss between these two seg-masks is calculated as is illustrated in Equ.~\ref{equ:RC}:
\begin{equation}
   \mathcal{L}_{drc}=-(S(W(x_{m},\hat{\varnothing}))-S(x_{f}))^{2}
   \label{equ:RC}
\end{equation}
where $x_{m}$, $x_{f}$ and $\hat{\varnothing_{n}}$ are the moving image, fixed image and deformation field from the DSS block. $W(\cdot,\cdot)$ is the deformation process in registration network and $S(\cdot)$ is the segmentation network. Each ROI is calculated in different channels obtaining independent fine optimization, while the task unconcerned regions are calculated in a background channel together. Thus, fine registration on ROIs is available and inter-regional error registration is avoided.
\subsection{Joint learning strategy exerts complementarity}
\label{sec2}
The registration network, segmentation network and pixel-wise discriminator in our DeepRS model (Fig.~\ref{fig:deeprs}) are trained by different loss function combinations to coordinate the training process and achieve mutual improvement.
\paragraph{Registration network} The registration network is optimized by four different targets. An adversarial loss $\mathcal{L}_{adv}$ \cite{fan2019adversarial} from the pixel-wise discriminator provides the similarity metric between warped and fixed images. The DRC loss $\mathcal{L}_{drc}$ from the segmentation network brings registration attention on ROIs. A local cross-correlation (CC) \cite{balakrishnan2018unsupervised} $\mathcal{L}_{cc}$ maintains the stability of the training process, and a smooth loss \cite{balakrishnan2018unsupervised} $\mathcal{L}_{R}$ penalizes local spatial variations in deformation field. Therefore, the total loss function $\mathcal{L}_{reg}$ is:
\begin{equation}
   \mathcal{L}_{reg}=\lambda_{adv}\mathcal{L}_{adv}+\lambda_{drc}\mathcal{L}_{drc}+\lambda_{cc}\mathcal{L}_{cc}+\lambda_{R}\mathcal{L}_{R}
   \label{equ:reg}
\end{equation}
\paragraph{Segmentation network} The loss function of the segmentation network $\mathcal{L}_{seg}$ consists of two components. One is the ACM loss $\mathcal{L}_{acm}$ that adds the weakly supervised data to the training for higher segmentation generalization ability. The other is cross-entropy loss $\mathcal{L}_{ce}$ between the warped images and labels that maintains the right optimization target:
\begin{equation}
   \mathcal{L}_{seg}=\lambda_{acm}\mathcal{L}_{acm}+\lambda_{ce}\mathcal{L}_{ce}
   \label{equ:seg}
\end{equation}
\paragraph{Pixel-wise discriminator} The training strategy of pixel-wise discriminator follows \cite{fan2019adversarial}: well-registered image pairs consisting of reference images $x_{r}$ and fixed images $x_{f}$ as positive cases and misaligned images consisting of warped images $x_{w}$ and fixed images $x_{f}$ as negative cases. The reference image $x_{r}$ is a fusion of a moving image $x_{m}$ and a fixed image $x_{f}$ according to the formula $x_{r}=\beta*x_{m}+(1-\beta)*x_{f}$. Thus, the loss for the discriminator $\mathcal{L}_{D}$ is:
\begin{equation}
   \mathcal{L}_{D}=-\log(D(x_{r},x_{f}))-\log(1-D(x_{w},x_{f}))
\label{equ:d}
\end{equation} 

\section{Experiments}
Extensive experimental results show that our DeepRS model enhances the performance of complex scene registration and few-shot segmentation tasks on cardiac CT data which has complex task-unconcerned backgrounds.
\subsection{Evaluation settings}
\paragraph{Dataset}
We validated the superiority of our DeepRS model on the whole heart registration and segmentation tasks on the CT dataset of \emph{MM-WHS 2017 Challenge} \cite{zhuang2016multi} which has complex backgrounds (lung, rib cage, etc.). This dataset consists of 20 labeled and 40 unlabeled CT images. Our experiments aim to register and segment seven cardiac structures including the ascending aorta, left atrial cavity (LA), left ventricular cavity(LV), myocardium of the left ventricle (Myo), pulmonary artery (PA), right atrial cavity (RA) and right ventricular cavity (RV). We first crop the rectangular regions containing the hearts for affine transformation and resample them to $128\times128\times96$. Then the labeled images are randomly split into 5 parts, 1 part (4 images) is used in training set as the moving images for few-shot situation and the remaining 4 parts (16 images) in testing set resulting in 5-folds evaluation. We put 40 unlabeled images as the fixed images in the training set leading to 160 data pairs together with the moving images, and the 16 images in the testing set are paired separately leading to 240 data pairs. Following the \cite{balakrishnan2018unsupervised}, we use Elastix\footnote{\url{https://www.elastix.org/}} to perform affine transformation so that our model only needs to pay attention to the deformation registration process.

\paragraph{Implementation}
The segmentation network and pixel-wise discriminator follow the same 3D U-Net \cite{cciccek20163d} structure. The registration network follows the VoxelMorph-2 \cite{balakrishnan2018unsupervised} structure. We use RMSprop \cite{tieleman2012lecture} to train the registration network and the discriminator for stable process \cite{arjovsky2017wasserstein}, and Adam \cite{kingma2014adam} to train the segmentation network for fast convergence. These models share the same learning rate of $2e^{-4}$ and training batch size of 1 due to the limitation of memory. According to extensive experiments, we finally set $\lambda_{adv}=1$, $\lambda_{drc}=10$, $\lambda_{cc}=1$, $\lambda_{R}=1$, $\lambda_{acm}=1$ and $\lambda_{ce}=1$. The models were implemented via Keras\footnote{\url{https://github.com/keras-team/keras}} with a Tensorlow\footnote{\url{https://github.com/tensorflow/tensorflow}} backend and were trained on a single NVIDIA TitanX GPU with 12 GB memory.

\paragraph{Comparison settings}
The comparison demonstrates the advancement on segmentation and registration of our DeepRS model. We compare our model's segmentation performance with three general segmentation networks (3D U-Net\cite{cciccek20163d}, V-Net\cite{milletari2016v}, 3D FCN\cite{long2015fully}) to illustrate the enhancement brought by registration. The 3D U-Net augmented by manual strategies (random rotate in $[-10^{\circ}, 10^{\circ}]$, random mirroring and random flipping) is compared with to show the advantages of registration-based data augmentation. We also compare two unsupervised registration models (VoxelMorph-2\cite{balakrishnan2018unsupervised}, Adv-Reg\cite{fan2019adversarial}) to illustrate the superiority of our deep-based region constraints in complex scene. In addition, two registration and segmentation joint models (DeepAtlas\cite{xu2019deepatlas}, HybridCNN\cite{li2019hybrid}) are compared with to demonstrate the superiority of the DeepRS brought by our DSS block, ACM method and DRC strategy. What's more, our proposed DeepRS is also evaluated on different data amount to illustrate its excellent generalization ability in few-shot segmentation. Finally, an ablation study is used to analyse the contributions of each our innovation.

\paragraph{Evaluation metric}
We evaluate the registration and segmentation methods with dice coefficient \cite{dice1945measures}. The dice coefficient ([\%]) is a metric that measures the coincidence degree between two sets according to $Dice(G,P)=\frac{2\mid G\bigcap P\mid}{\mid G\mid+\mid P\mid}$ where the $G$ is the ground truth and the $P$ is the predicted mask. It is suitable to evaluate the agreement between the predicted segmentation/registration and the ground truth. The Dice coefficients of the corresponding seven cardiac structures are calculated, and presented as $mean\pm std$.
\subsection{Results}
Extensive experimental results on cardiac CT dataset show that with merely 4 training labels, our proposed DeepRS appears to be a strong superiority both in the quantitative comparison and in visual. The experiments on different label amounts illustrate that our DeepRS greatly reduces the label dependence of the segmentation model.
\subsubsection{Quantitative comparison}
\begin{figure*}
\begin{center}
\includegraphics[width=11cm]{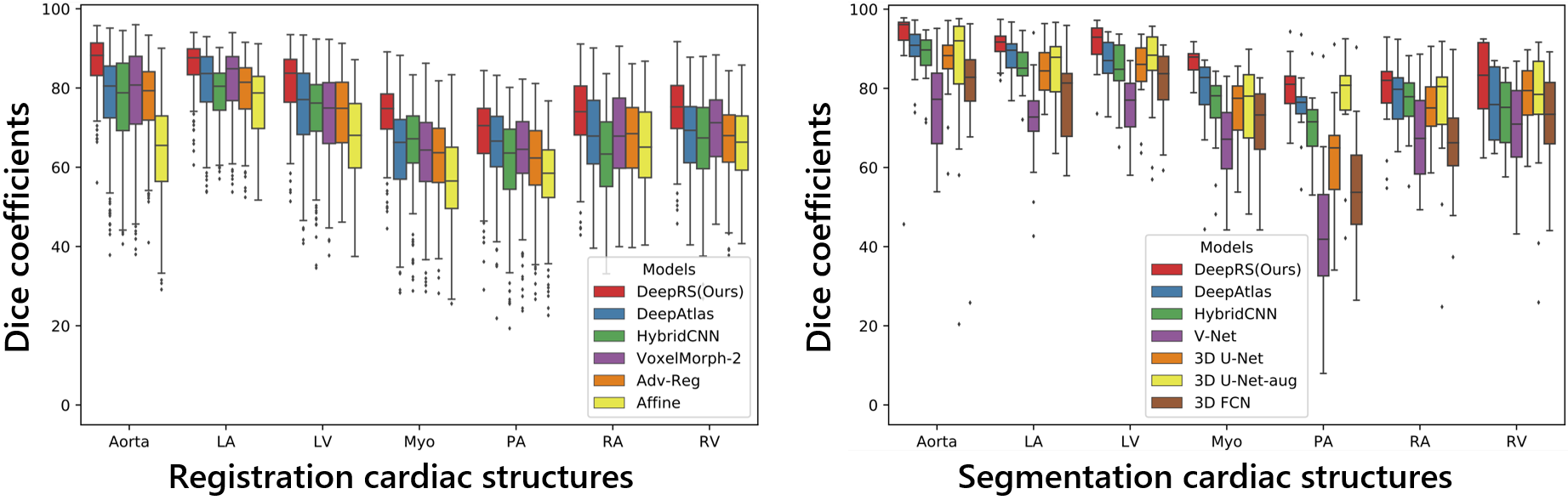}
\end{center}
   \caption{Our DeepRS achieves excellent dice coefficients on each structure. The box plots shows the proposed DeepRS (red box) model achieves the state-of-the-art performance in complex scene registration (\emph{Left}) and few-shot segmentation (\emph{Right}).}
\label{fig:reg_seg}
\end{figure*}
As shown in Tab.~\ref{tab:metrics}, our DeepRS model achieves the state-of-the-art performance in both registration and segmentation tasks whose mean dice coefficients of all cardiac structures are 77.6\% and 85.7\%. Fig.~\ref{fig:reg_seg} illustrates that the proposed DeepRS achieves excellent dice coefficients on each structure in complex scene registration and few-shot segmentation.
\begin{wraptable}{r}{7cm}
\centering
\caption{The proposed DeepRS model achieves the state-of-the-art performance both in registration (R) and segmentation (S) tasks on cardiac CT data.}
\begin{tabular}{ccc}
\hline
Method                                          &R-Dice                 &S-Dice\\
\hline
Affine only                                     &64.6$\pm$10.7          &-\\
VoxelMorph-2\cite{balakrishnan2018unsupervised} &71.7$\pm$10.6          &-\\
Adv-Reg\cite{fan2019adversarial}                &68.8$\pm$10.7           &-\\
\hline
3D U-Net\cite{cciccek20163d}                    &-                      &78.8$\pm$9.2\\
3D U-Net-aug\cite{cciccek20163d}                &-                      &80.0$\pm$12.0\\
3D FCN\cite{long2015fully}                      &-                      &71.4$\pm$11.3\\
V-Net\cite{milletari2016v}                      &-                      &69.8$\pm$10.9\\
\hline
DeepAtlas\cite{xu2019deepatlas}                 &71.3$\pm$10.5          &81.8$\pm$7.5\\
HybridCNN\cite{li2019hybrid}                    &69.2$\pm$10.3          &78.8$\pm$7.9\\
\hline
\textbf{DeepRS(Ours)}                           &\textbf{77.6$\pm$7.9}  &\textbf{85.7$\pm$7.7}\\
\hline
\end{tabular}
\label{tab:metrics}
\end{wraptable}

On the registration task, the registration network gets the deep-based region constraints from the segmentation network, bringing finer registration on ROIs in complex scene than other registration models. VoxelMorph-2 lacks region constraints, thus the dice is 5.9\% lower than ours. The Adv-Reg takes a GAN whose training process is unstable to learn a similarity metric and gets worse results than VoxelMorph-2. DeepAtlas utilizes weakly supervised data directly, thus the misaligned regions disturbs the training process of the segmentation model finally in turn disturbing the registration performance (71.3\%). HybridCNN lacks label-based region constraints in our few-shot situation thus the influence of the misaligned regions are more pronounced (69.2\%).

On the segmentation task, the segmentation network in our DeepRS model effectively utilizes the augmentation data and weakly supervised data from the registration network, thus achieving much higher dice coefficient than 3D U-Net, 3D FCN and V-Net. Although the 3D U-Net augmented by manual strategies has get 1.2\% dice improvement compared with non-augmentation, our DeepRS model has even greater advantage by 5.7\%. Due to the influence of the misaligned regions in weakly supervised data, the HybridCNN only gets 78.8\% dice. Similarly, the DeepAtlas takes the augmentation data from registration, but the misaligned regions still limits the enhancement which make it get only 81.8\% dice.

\subsubsection{Visual superiority}
Visually, our DeepRS model brings higher segmentation generalization ability with few labels, and achieves finer registration performance on ROIs in complex scene.

\begin{wrapfigure}{r}{0.6\linewidth}
\begin{center}
\includegraphics[width=\linewidth]{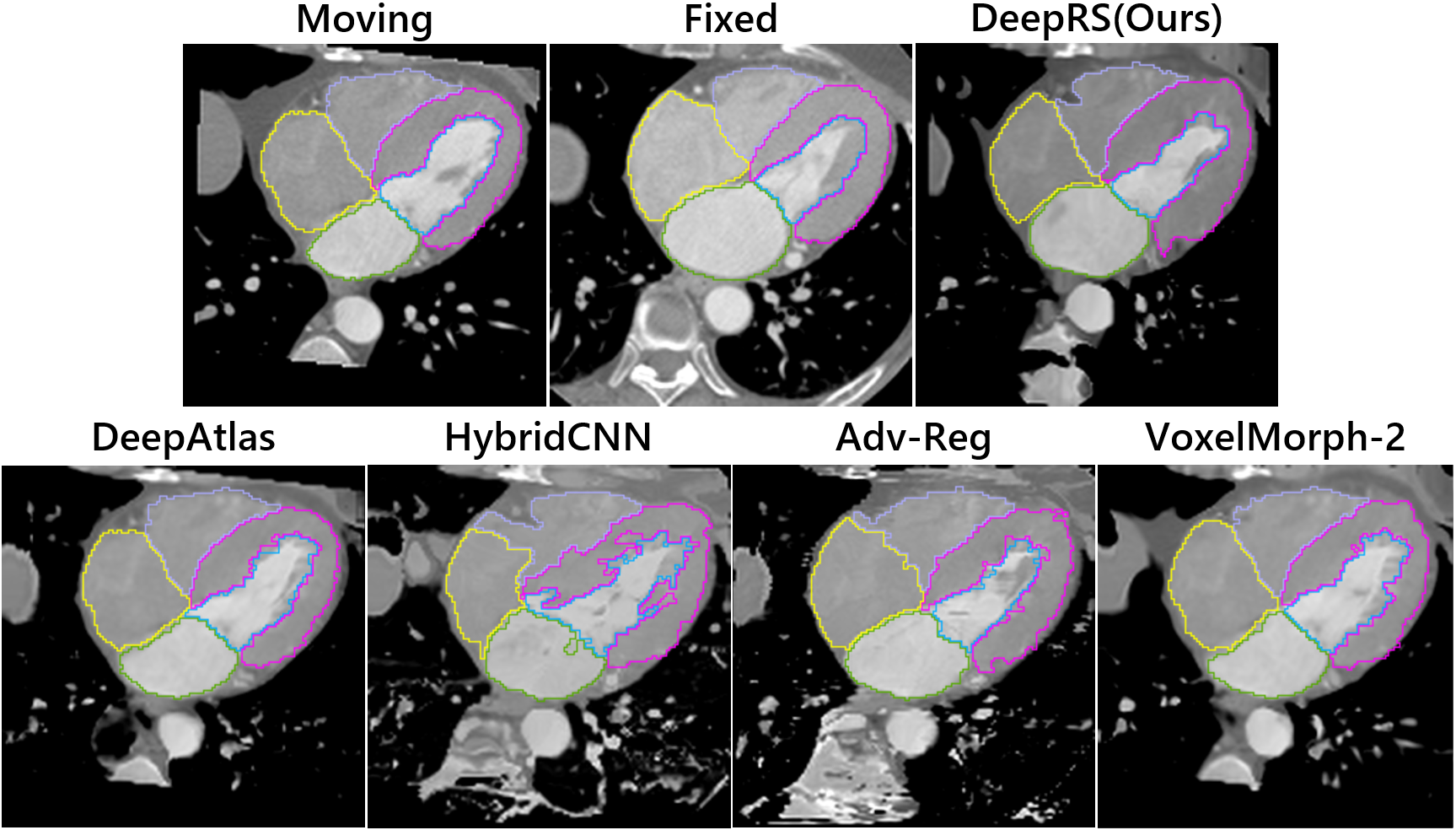}
\end{center}
   \caption{Our DeepRS gets finer registration on ROIs. The example slices from 3D CT image show the overlaid boundaries of the LV (green), RA (yellow), RV (purple), LV (blue) and Myo (pink). Our model makes these structures in moving image alike structures in fixed image.}
\label{fig:visual_reg}
\end{wrapfigure}

As illustrated in Fig.~\ref{fig:visual_reg}, our DeepRS brings finer registration on ROIs making 5 structures in moving image look more similar to these structures in fixed image. The HybridCNN uses weakly supervised data directly and lacks label-based region constraints. Therefore the misaligned regions interrupt the segmentation training process and in turn weaken the registration performance bringing serious region correspondence errors. The Adv-Reg is optimized by the unstable GAN making the warped image messy and rough in detail.

\begin{figure*}
\begin{center}
\includegraphics[width=\linewidth]{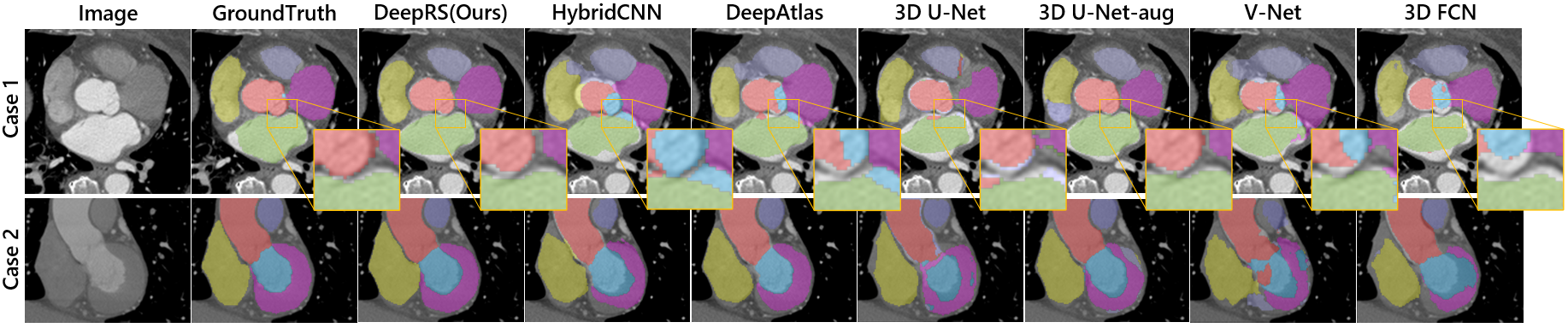}
\end{center}
   \caption{Our DeepRS brings higher segmentation generalization ability trained on 4 labeled images. Yellow boxes show the excellent generalization ability in detail. The example slices from 3D CT image show the regions of Aorta (red), RA (yellow), RV (purple), Myo (pink), LV (green) and LV (blue).}
\label{fig:visual_seg}
\end{figure*}

As shown in Fig.~\ref{fig:visual_seg}, our DeepRS model brings much higher segmentation generalization ability trained on merely 4 labeled images. Case 1 shows the excellent generalization ability and the yellow boxes show the performance in detail. Our DeepRS has achieved fine segmentation, while the 3D U-Net, 3D FCN, 3D U-Net-aug and V-Net have many mis-segmentation regions. The HybridCNN and DeepAtlas has more mis-segmentation regions than others due to the misaligned regions in weakly supervised data. Case 2 shows the fine segmentation capability in another perspective and sample. The 3D U-Net ,3D FCN and V-Net are limited by small dataset leading to various serious mis-segmentation.
\subsubsection{DeepRS for few-shot segmentation}
\begin{wrapfigure}{r}{0.6\linewidth}
\begin{center}
\includegraphics[width=\linewidth]{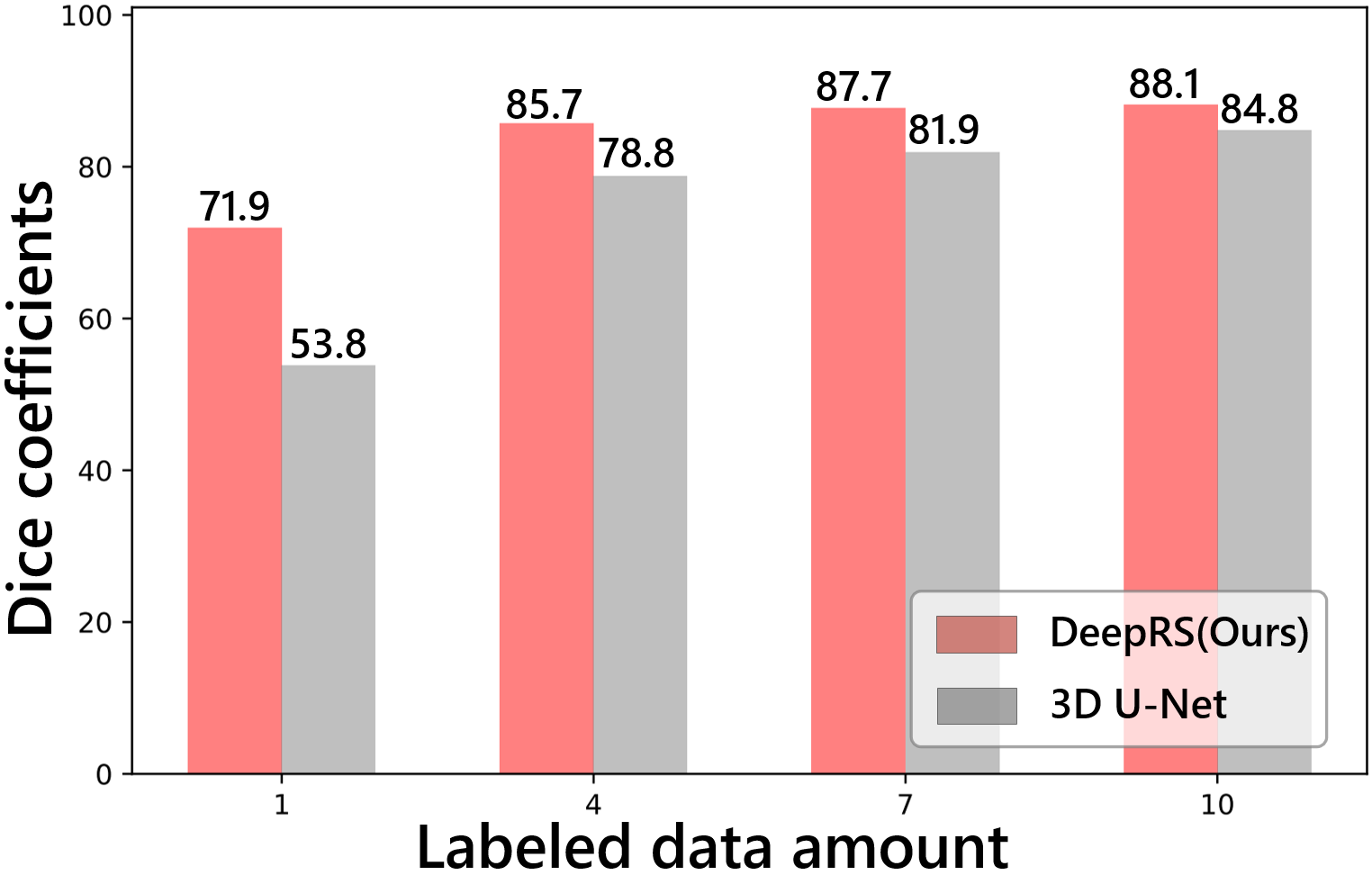}
\end{center}
   \caption{Especially in few-shot situation, the segmentation network in our DeepRS model achieves much higher mean dice coefficients of all structures than 3D U-Net\cite{cciccek20163d}.}
\label{fig:fewshot}
\end{wrapfigure}

In few-shot situation, the segmentation (S) network in our DeepRS model achieves higher mean dice coefficients of all structures than 3D U-Net as illustrated in Fig.~\ref{fig:fewshot}. The effectiveness of our DeepRS is evaluated on randomly-sampled labeled data whose amount is 1, 4, 7 and 10 respectively. 3D U-Net is used for comparison and the mean dice coefficients of all structures are calculated. As the labeled data decreases, the superiority of our DeepRS on segmentation task becomes more prominent. When only one label is available, our segmentation performance is 18.1\% higher than 3D U-Net.
\subsection{Ablation study}
\begin{wraptable}{r}{7cm}
\centering
\caption{The ablation study analyses the contributions of our innovations.}
\begin{tabular}{ccccc|cc}
\hline
R           &S           &DSS         &ACM              &DRC         &R-Dice               & S-Dice\\
\hline
\checkmark  &            &            &                 &            &72.2$\pm$10.3        &-\\
            &\checkmark  &            &                 &            &-                    &78.8$\pm$9.2\\
\checkmark  &\checkmark  &            &                 &            &72.9$\pm$10.4        &80.5$\pm$10.2\\
\checkmark  &\checkmark  &\checkmark  &                 &            &72.9$\pm$9.6         &83.9$\pm$8.3\\
\checkmark  &\checkmark  &            &\checkmark       &            &72.5$\pm$10.1        &84.1$\pm$8.3\\
\checkmark  &\checkmark  &            &                 &\checkmark  &75.9$\pm$9.1         &82.5$\pm$9.2\\
\checkmark  &\checkmark  &\checkmark  &\checkmark       &\checkmark  &\textbf{77.6$\pm$7.9}  &\textbf{85.7$\pm$7.7}\\
\hline
\end{tabular}
\label{tab:ablation}
\end{wraptable}
As shown in Tab.~\ref{tab:ablation}, an ablation study illustrates each great advantage brought by our innovations. The directly joint model only utilizes the registration's data augmentation ability thus the segmentation gets 80.5\% dice and the registration gets 72.9\% dice. Our DSS block embeds a random perturbation factor in the registration to maintain the diversity of augmentation data (\textbf{Solution 1}), thus bringing 3.4\% segmentation dice growth. The ACM method adds the supervision information in weakly supervised data (\textbf{Solution 2}) to segmentation network so that it gets 3.6\% segmentation dice improvement. The DRC strategy builds deep-based region constraints instead of label-based methods (\textbf{Solution 3}) via the warped and fixed segmentations increasing the direct joint model by 3\% registration dice. We find that the segmentation and registration models achieve further promotion in our final DeepRS model owing to their complementarity, thus finally achieving 77.6\% registration dice and 85.7\% segmentation dice which are increased by 4.7\% and 5.2\% respectively.

\section{Conclusion}
This paper presents a \emph{Deep Complementary Joint Model(DeepRS)} for complex scene registration and few-shot segmentation. Our proposed \emph{DSS block} adjusts deformation fields randomly via a perturbation factor, thus increasing the activity of the warped images and labels and achieving sustainable data augmentation capability. Our proposed \emph{ACM method} efficiently utilizes the supervision information in weakly supervised data via alignment confidence maps from a pixel-wise discriminator bringing higher segmentation generalization. Our proposed \emph{DRC strategy} constructs label-free loss between the warp and fixed images from the segmentation model resulting in finer registration on ROIs. We train our proposed DeepRS model on the cardiac CT dataset which has complex background with few labels with merely 4 labels and shows great advantages in registration and segmentation tasks compared to existing methods.

Our work greatly reduces the requirement of a large labeled dataset and provides the fine optimization targets, thus the registration and segmentation accuracy are improved and the cost is greatly saved. Especially, our DeepRS model has great potential in some situations where the labeling is difficult, the scene is complex or the dataset is small. 

\subsubsection*{Acknowledgments}
This research was supported by the National Natural Science Foundation under grants (61828101,31571001,31800825), the Short-Term Recruitment Program of Foreign Experts (WQ20163200398),  and Southeast University-Nanjing Medical University Cooperative Research Project (2242019K3DN08). We thank the Big Data Computing Center of Southeast University for providing the facility support on the numerical calculations in this paper.
%
%
\bibliographystyle{splncs04}
\bibliography{egbib}
\end{document}